\documentclass[10pt,twocolumn,letterpaper]{article}

\usepackage{cvpr}              
\usepackage{algorithm}
\usepackage{algorithmic}
\usepackage{amssymb}
\usepackage{amsmath}
\usepackage{multirow}
\usepackage{booktabs}
\usepackage{color}

\definecolor{cvprblue}{rgb}{0.21,0.49,0.74}
\usepackage[pagebackref,breaklinks,colorlinks,allcolors=cvprblue]{hyperref}

\def\paperID{9316} 
\def\confName{CVPR}
\def\confYear{2025}

\title{CountLLM: Towards Generalizable Repetitive Action Counting\\via Large Language Model}

\author{Ziyu Yao$^{1}$ \quad Xuxin Cheng$^{1}$ \quad Zhiqi Huang$^{1}$ \quad Lei Li$^{2,3}$\footnotemark[1] \\
$^1$Peking University \quad
$^2$University of Washington \quad
$^3$University of Copenhagen \\
{\tt \small \{yaozy, chengxx\}@stu.pku.edu.cn} \quad {\tt \small zhiqihuang@pku.edu.cn} \quad {\tt \small lilei@di.ku.dk} \\
}

\begin{document}
\maketitle

\renewcommand{\thefootnote}{\fnsymbol{footnote}}
\footnotetext[1]{Corresponding author.}

\begin{abstract}
Repetitive action counting, which aims to count periodic movements in a video, is valuable for video analysis applications such as fitness monitoring. However, existing methods largely rely on regression networks with limited representational capacity, which hampers their ability to accurately capture variable periodic patterns. Additionally, their supervised learning on narrow, limited training sets leads to overfitting and restricts their ability to generalize across diverse scenarios.
To address these challenges, we propose \textbf{CountLLM}, the first large language model (LLM)-based framework that takes video data and periodic text prompts as inputs and outputs the desired counting value.
CountLLM leverages the rich clues from explicit textual instructions and the powerful representational capabilities of pre-trained LLMs for repetitive action counting.
To effectively guide CountLLM, we develop a periodicity-based structured template for instructions that describes the properties of periodicity and implements a standardized answer format to ensure consistency.
Additionally, we propose a progressive multimodal training paradigm to enhance the periodicity-awareness of the LLM.
Empirical evaluations on widely recognized benchmarks demonstrate CountLLM's superior performance and generalization, particularly in handling novel and out-of-domain actions that deviate significantly from the training data, offering a promising avenue for repetitive action counting.
\end{abstract}

\section{Introduction}

With the rapid advancements in video understanding~\cite{liu2022video, wang2023videomae, wang2022internvideo, wang2024internvideo2, li2022uniformerv2,an2023temporal,li2023unmasked,li2024human}, repetitive action counting (RAC) has become a critical area of study. This task, which involves counting repetitive activities in videos, plays a crucial role in analyzing periodic actions, such as those in fitness monitoring~\cite{fieraru2021aifit}, pedestrian detection~\cite{ran2007pedestrian}, camera calibration~\cite{huang2016camera}, and 3D reconstruction~\cite{li2018structure, ribnick20103d}.

\begin{figure}[t!]
  \includegraphics[width=\linewidth]{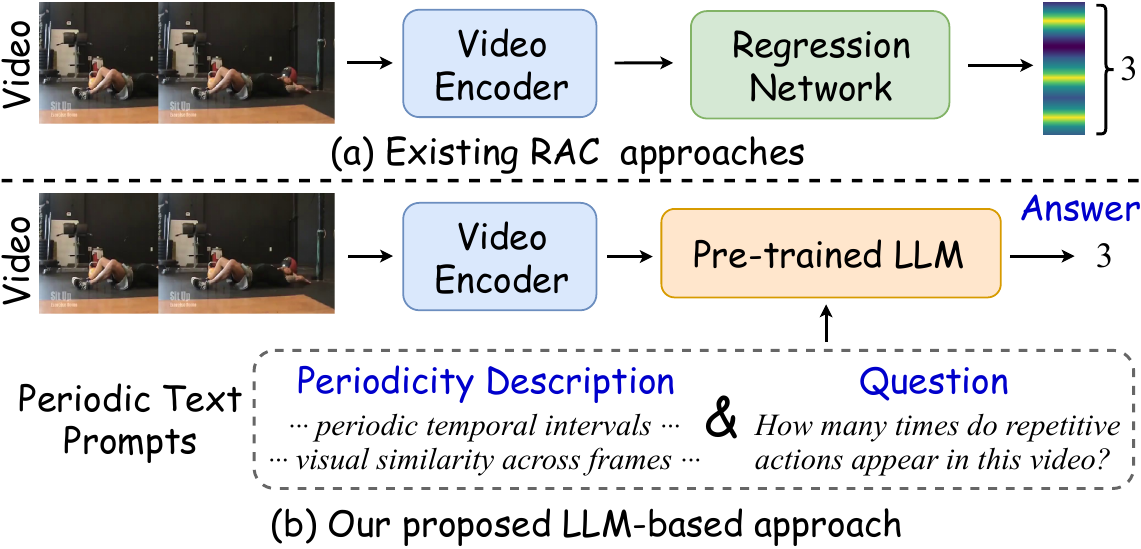}
  \caption{\textbf{a):} Existing approaches directly regress the density map, and sum the values of each frame to obtain the final count.  
  \textbf{b):} Our proposed LLM-based approach utilizes explicit periodic text prompts to describe key properties of periodicity, and leverages the powerful reasoning capabilities of pre-trained LLMs, achieving accurate and generalizable repetitive action counting.}
  \label{fig:introduction}
  \vspace{-0.1cm}
\end{figure}

Despite its importance, this field remains underexplored. Previous approaches mostly adopt a straightforward strategy: first encoding video features, then regressing density maps, and finally summing the values to achieve counting~\cite{hu2022transrac, sinha2024every, dwibedi2024ovr} (as shown in~\cref{fig:introduction} (a)).
However, they exhibit significant limitations in addressing the diversity and complexity of the RAC task. First, the cycle lengths of different actions vary greatly and are unevenly distributed; moreover, videos often contain complex backgrounds and motion blur, which interfere with accurate counting. Existing approaches largely rely on manually designed regression neural networks, whose representational capacity is limited, making it difficult to accurately capture such variable and non-homogeneous periodic patterns. Second, the current research community's definition of actions is overly narrow, as the actions in existing datasets are limited and primarily focus on human fitness activities (such as squats, crunches, and jumping jacks). This results in traditional supervised learning models being prone to overfitting when trained on training sets of these small datasets, thus limiting their ability to generalize to out-of-domain scenarios.

To address these challenges, we propose integrating large-scale pre-trained large language models (LLMs) into the RAC task to leverage their powerful representational and generalization capabilities. Due to their pre-training on massive real-world data, LLMs possess strong representational abilities. As illustrated in~\cref{fig:introduction} (b), by replacing the regression network in existing approaches, LLMs can effectively capture actions and their periodic patterns, maintaining high counting accuracy even in the presence of complex backgrounds or motion blur. Furthermore, through our explicit periodic text prompts combined with instruction fine-tuning scheme, LLMs gain enhanced periodicity awareness, enabling them to generalize not only to human activities but also to a broader range of actions, such as animal movements (\eg, cats, dogs) or periodic motions of non-living entities (\eg, clock movements, planetary motions).

Building on these insights, we present \textbf{CountLLM}, the first LLM-based approach for repetitive action counting. CountLLM formulates RAC as a video question-answering problem, where both video data and periodic text prompts are used as inputs to generate accurate counts. The framework consists of a pre-trained video encoder, a periodicity transformer, a linear projector, and a pre-trained LLM. The video encoder extracts video features, from which the periodicity transformer derives relevant periodic representations. These representations are then converted into periodic tokens by the linear projector, subsequently combined with textual tokens as inputs to the LLM.

To effectively guide our CountLLM in understanding periodicity, we carefully develop a structured template for instruction-driven conversation, specifically tailored for the RAC task. The template encompasses periodicity description, question, and answer. Here, the description includes the main properties of periodicity, such as periodic temporal intervals and visual similarities across frames, which assist the LLM in capturing periodic patterns. Meanwhile, the answer follows a standardized format to ensure consistency, especially for long videos that require sequential splitting and processing. Additionally, we implement a progressive multimodal training strategy incorporating \textit{Periodicity-aware Alignment} to enhance the CountLLM's ability to interpret and extract periodic patterns.

We summarize our contributions in three-fold:
\begin{itemize}
    \item To the best of our knowledge, we are the first to apply LLMs to the RAC task and introduce \textbf{CountLLM}, which utilizes the rich clues from explicit text prompts and leverages the powerful capability of pre-trained LLMs.

    \item We present a structured template for instruction-driven conversations in the context of RAC and design a progressive training paradigm to optimize CountLLM, thereby enhancing its periodicity awareness.

    \item CountLLM achieves state-of-the-art performance on three widely recognized benchmarks and demonstrates excellent generalization in out-of-domain scenario, providing a promising approach for action counting.
\end{itemize}

\section{Related Works}

\paragraph{Repetitive Action Counting} Early methods~\cite{chetverikov2006motion,laptev2005periodic,lu2004repetitive} compress the motion field into 1D signals via Fourier~\cite{pogalin2008visual}, peak~\cite{thangali2005periodic}, and wavelet~\cite{runia2018real}, but they assume uniform periodicity and struggle with non-stationary repetitions.

Recent advances in deep learning have led to improved approaches.
For example, Context~\cite{zhang2020context} uses a context-aware regression network with coarse-to-fine refinement for counting.
RepNet~\cite{dwibedi2020counting} constructs temporal self-similarity matrices based on pair-wise similarity of embeddings.
TransRAC~\cite{hu2022transrac} encodes multi-scale temporal correlations to handle various action frequencies.
ESCounts~\cite{sinha2024every} introduces an exemplar-based encoder-decoder framework to improve counting accuracy.
Another line of work integrates other modalities, including audio~\cite{zhang2021repetitive}, human pose~\cite{yao2023poserac}, and optical flow~\cite{li2024repetitive}, to enhance counting.

However, previous methods mainly relied on manually designed regression neural networks and supervised learning on limited datasets, restricting their representational capacity and generalizability. In contrast, our approach leverages explicit text prompts and powerful pre-trained LLMs, making it more generalizable for counting novel actions.

\vspace{-0.3cm}

\begin{figure*}[t!]
  \includegraphics[width=\linewidth]{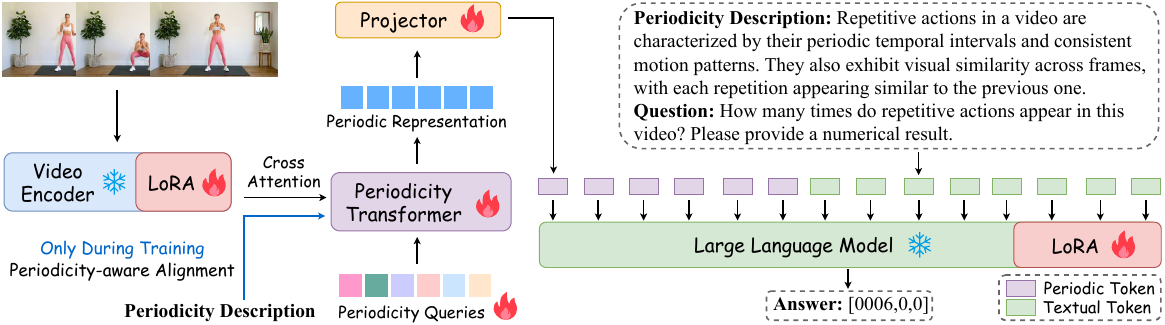}
  \caption{Our proposed CountLLM takes video and periodic text prompt as input and consists of four components: a pre-trained video encoder, a periodicity transformer, a projector, and a pre-trained LLM. Guided by our proposed Periodicity-aware Alignment, the periodicity transformer captures periodic representations from the video features. These representations are then projected into periodic tokens by the projector. The pre-trained LLM takes both the periodic and textual tokens as input, and outputs the corresponding counting values.}  
  \label{fig:framework}
\end{figure*}

\paragraph{Multimodal Large Language Model} Based on large language models (LLMs)~\cite{brown2020language, touvron2023llama}, studies are increasingly exploring their adaptation to other modalities, such as image~\cite{chen2024internvl,cai2025bayesian, han2024onellm, lu2024unified,li2024cpseg}, point cloud~\cite{qi2024gpt4point, xu2023pointllm,an2024multimodality,an2024rethinking}, and audio~\cite{huang2024audiogpt, zhang2023speechgpt,yang2025you}, to develop multimodal LLMs (MLLMs). In the video domain, numerous video MLLM works~\cite{ye2024mplug, zhang2023video, li2023videochat, lin2023video, xu2024slowfast, song2024moviechat, yu2024self, huang2024vtimellm, ren2024timechat, qu2024llms,lei2025chatmotion} have emerged.
Among these, VideoChatGPT~\cite{maaz2023video} applies mean pooling on frame encodings before feeding them into LLMs.
Chat-UniVi~\cite{jin2024chat} employs DPC-KNN to cluster dynamic visual tokens for multi-scale features.
VideoChat2~\cite{li2024mvbench} uses a Q-Former~\cite{li2023blip} to connect with LLMs for temporal understanding.
ST-LLM~\cite{liu2024st} introduces dynamic masking and a global-local input module to enhance modeling.

However, existing video MLLMs focus on general video understanding, lacking specific awareness of periodicity. We are the first to leverage LLMs for the RAC task, demonstrating that LLMs can effectively achieve superior performance in generalizable repetitive action counting.
\section{Methodology}

\subsection{Problem Formulation}

Given a video $\mathcal{V}={\{x_i\}}^{T}_{1}\in \mathbb{R}^{3\times H\times W\times T}$ with $T$ frames, RAC model aims to predict a value $\mathcal{M}$, which is the number of repetitive actions. This task can be formulated as:
\begin{equation}
\label{eq:form1}
\mathcal{M} = \operatorname{counting}(\mathcal{V}).
\end{equation}
Following~\cref{eq:form1}, previous approaches~\cite{hu2022transrac, zhang2020context, dwibedi2020counting} rely on manually designed regression networks to perform supervised learning from the training set, which limits their generalization ability for unseen and out-of-domain actions.

In contrast to the previous formulation, we investigate the concept of periodicity and incorporate its main properties into explicit text prompts. We then leverage the powerful LLMs to achieve more accurate and generalizable repetitive action counting. We reformulate~\cref{eq:form1} as follows:
\begin{equation} \label{eq:form2} 
\mathcal{M} = \operatorname{counting}(\mathcal{V}, \mathcal{T}), 
\end{equation}
where $\mathcal{T}$ represents the given periodic text prompts. Based on~\cref{eq:form2}, we frame the RAC task as a video question-answering (VQA) task and utilize the powerful reasoning capability of LLMs to perform action counting.

\subsection{Overview}

As illustrated in~\cref{fig:framework}, we propose the first LLM-based repetitive action counting approach, named \textbf{CountLLM}. CountLLM consists of four components: a video encoder $\Phi_{V}(\cdot)$, a periodicity transformer $\Phi_{F}(\cdot)$, a linear projector $\Phi_{P}(\cdot)$, and a large language model $\Phi_{L}(\cdot)$. The input includes two parts: the video $\mathcal{V}\in \mathbb{R}^{3\times H\times W\times T}$ and the periodic text prompt $\mathcal{T}$, and the output is the corresponding number of repetitive actions $\mathcal{M}$. Based on these symbols, we define the overall pipeline as follows:
\begin{equation}
\mathcal{M} = \Phi_{L}(\Phi_P(\Phi_F(\Phi_V(\mathcal{V}))), \mathcal{T}).
\end{equation}

Initially, the video encoder $\Phi_{V}(\cdot)$ processes a video clip $\mathcal{V} \in \mathbb{R}^{3\times H\times W\times T}$ to generate video features $\mathcal{F} = {\{f_i\}}_{i=1}^{m} \in \mathbb{R}^{D_V \times m}$, where $m$ and $D_V$ denote the number and dimension of these features, respectively.

To refine and focus on critical periodic representations within the video, we introduce a periodicity transformer that compresses the redundant elements of $\mathcal{F}$, capturing valuable periodic structures that serve as context for the LLM. Specifically, we define a fixed number of learnable periodicity queries $\mathcal{Z}^0 \in \mathbb{R}^{D_Z \times n}$, where $n$ and $D_Z$ represent the number and dimension of the queries, respectively. $\mathcal{Z}^0$ interacts with $\mathcal{F}$ through cross-attention layers to focus on core periodic information while disregarding other factors, such as background and motion blur. After passing through an $L$-layer periodicity transformer, we obtain periodic representations $\mathcal{Z}^L \in \mathbb{R}^{D_Z \times n}$. Finally, $\mathcal{Z}^L$ is projected into periodic tokens $\mathcal{P}_{\mathrm{token}}$ by a projector $\Phi_{P}(\cdot)$:
\begin{equation}
\mathcal{P}_{\mathrm{token}} = {\{p_i\}}_{i=1}^{n} = \Phi_P(\Phi_F(\Phi_V(\mathcal{V}))),
\end{equation}
where ${\{p_i\}}_{i=1}^{n} \in \mathbb{R}^{D_L \times n}$ represents $n$ periodic tokens, each with a dimension of $D_L$, matching the LLM.

Meanwhile, the periodic text prompt $\mathcal{T}$ is processed by the tokenizer of $\Phi_L(\cdot)$ to produce a sequence of $s$ tokens, denoted as $\mathcal{T}_{\mathrm{token}} = {\{t_i\}}_{i=1}^{s}$. We combine the periodic tokens $\mathcal{P}_{\mathrm{token}}$ and the textual tokens $\mathcal{T}_{\mathrm{token}}$ as input to the LLM. Through our designed instruction fine-tuning scheme (which will be detailed later), the LLM gains enhanced periodicity awareness, enabling it to understand the contextual relationships between these tokens and generate accurate responses. Suppose the output is $\mathcal{R}_{\mathrm{token}} = {\{r_i\}}_{i=1}^{l}$ with $l$ tokens, this process can be defined as follows:
\begin{equation}
\{r_1, r_2, \dots, r_l\}=\Phi_L(\{p_1, \dots, p_n, t_1, \dots, t_s\}),
\end{equation}
where each $r_i$ is generated sequentially, taking into account all preceding tokens $\{p_1, \dots, p_n, t_1, \dots, t_s, r_1, \dots, r_{i-1}\}$. Finally, each output token $r_i$ is mapped to the LLM vocabulary via a linear classifier $\mathcal{C}(r_i)$, and the word with the highest probability is chosen.

To facilitate the repetitive action counting task, the design of both the instruction conversation and the optimization strategy is critical in enhancing CountLLM's ability to accurately model and respond to repetitive actions in video sequences. Therefore, the following two sections proceed to detail these aspects, respectively.

\subsection{Instruction Design}

Designing instructions is crucial for tuning LLMs to specific tasks. As shown in~\cref{fig:framework}, we develop a structured template for instruction-driven conversation tailored for the RAC task: ``\{\texttt{periodic tokens}\}, \{\texttt{periodicity description}\}, \{\texttt{question}\}, \{\texttt{answer}\}''. Details on the description and answer formats are provided below.

\vspace{-0.1cm}

\paragraph{Periodicity Description}
Before questioning the LLMs, we provide a detailed description of repetitive actions in a video to help the LLMs understand this challenging task. Specifically, we prompt GPT-4~\cite{achiam2023gpt} to describe the characteristics of repetitive or periodic actions in the video, ensuring that the textual description outlines the main properties of periodicity. Additionally, we emphasize that these actions should be general and class-agnostic. We then manually review and refine the GPT-4's responses to ensure accuracy and conciseness. The final description is shown in~\cref{fig:framework}.

\vspace{-0.1cm}

\paragraph{Answer Format} 

As our approach utilizes the LLM for action counting, the output is in text format. Therefore, the challenge lies in formatting the counting values such that the LLM can predict them accurately. To address this, we standardize the LLM's output format as $[abcd, e, f]$.

In this format, $abcd$ is a decimal string representing the number of repetitive actions, where each of $a$ to $d$ denotes a digit between $0$ and $9$. As a result, $abcd$ can represent any integer between $0000$ and $9999$. The decimal string $abcd$ has a fixed length and is prefixed with zeros if necessary.

\begin{figure}[t!]
  \includegraphics[width=\linewidth]{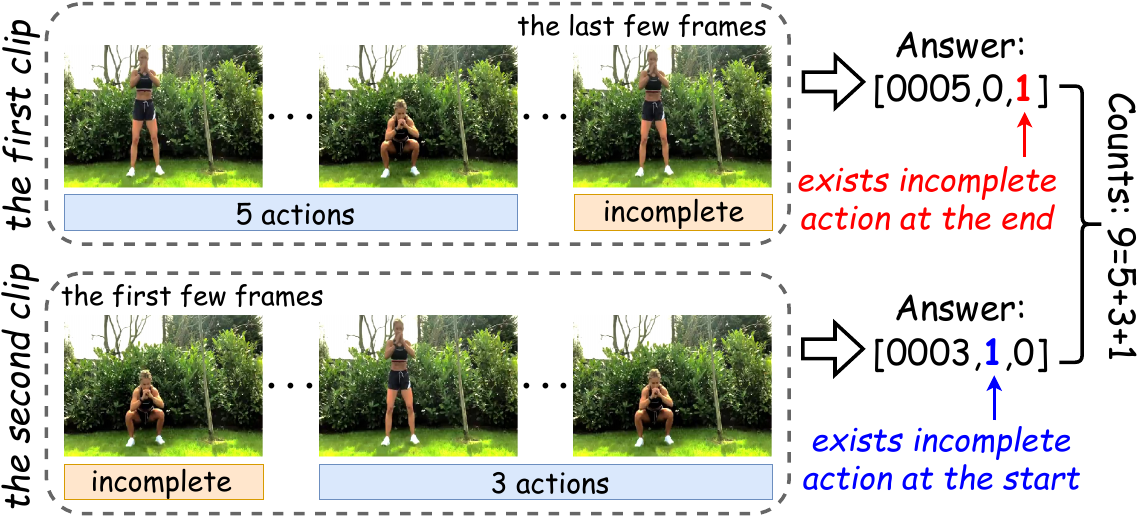}
  \caption{Our standardized answer format is $[$\textcolor[RGB]{255, 128, 0}{$\boldsymbol{abcd}$}$,$\textcolor[RGB]{0, 0, 255}{$\boldsymbol{e}$}$,$\textcolor[RGB]{255, 0, 0}{$\boldsymbol{f}$}$]$, where the decimal string \textcolor[RGB]{255, 128, 0}{$\boldsymbol{abcd}$} represents the counting value within this video clip, and \textcolor[RGB]{0, 0, 255}{$\boldsymbol{e}$} and \textcolor[RGB]{255, 0, 0}{$\boldsymbol{f}$} indicate whether there is an incomplete action cycle at the \textcolor[RGB]{0, 0, 255}{start} and \textcolor[RGB]{255, 0, 0}{end} positions, respectively (a value of $1$ denotes an incomplete action cycle). A complete action cycle is considered split across two clips only if both \textcolor[RGB]{255, 0, 0}{$\boldsymbol{f}$} in the last clip and \textcolor[RGB]{0, 0, 255}{$\boldsymbol{e}$} in the subsequent clip are $1$. In this case, the video-level counting value should increase by one (as shown in the figure).}
  \label{fig:answer}
\end{figure}

Meanwhile, we also predict $e$ and $f$ as flags to indicate incomplete action cycles at the start and end positions of each clip, respectively. Specifically, we set $e$ and $f$ to $1$ if an incomplete action cycle is present at the start or end of a clip and to $0$ otherwise. This approach addresses challenges in counting actions within long videos, where a common method is to split the video into multiple clips and input them sequentially into the LLM. However, such splits may result in action cycles being divided across two clips, as illustrated in~\cref{fig:answer}. In these cases, we increment the video-level counting value by $1$ only if both $f$ in the preceding clip and $e$ in the following clip are $1$, indicating a complete action cycle that spans two clips.

Overall, our proposed answer format has two advantages:
\textbf{1)} The answer is in decimal string format, meaning that all tokens are already included in the LLM vocabulary, eliminating the need to add new learnable tokens.
\textbf{2)} This format can effectively leverage the reasoning ability of the LLM to handle action counting for overly long videos.

\begin{figure*}[t!]
    \centering
    \includegraphics[width=0.95\linewidth]{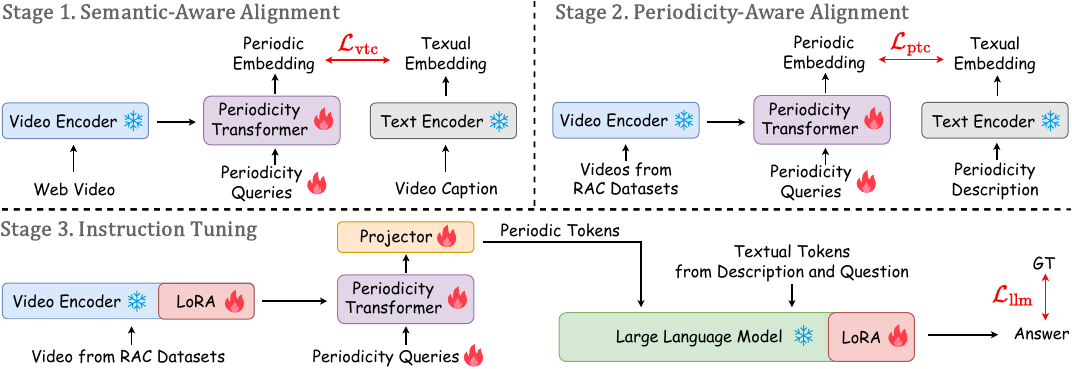}
    \caption{Progressive multimodal training paradigm of our CountLLM. In stage 1, periodic embeddings are aligned with video captions to establish text semantic awareness, bridging the multimodal gap. Based on this semantic understanding, stage 2 further aligns periodic embeddings with textual descriptions of periodicity, enabling a seamless transition to periodicity-awareness. Finally, stage 3 conducts parameter-efficient instruction tuning, enhancing the LLM's understanding and focus on the RAC task.}
    \label{fig:training}
\end{figure*}

\subsection{Progressive Multimodal Training}
To effectively enhance CountLLM's awareness of periodic patterns in video, we propose a progressive multimodal training paradigm consisting of three stages, as illustrated in~\cref{fig:training}. In the first two stages, we progressively optimize the periodicity transformer to extract robust periodic representations. In stage 1, we align its embeddings with video captions to achieve text semantic awareness, followed by alignment with textual periodicity description in stage 2. Building on the semantic understanding acquired in stage 1, the periodicity transformer interprets periodic descriptions in stage 2, facilitating a smooth transition to periodicity-awareness. Finally, in stage 3, we optimize the entire CountLLM through instruction tuning.

\paragraph{Stage 1. Semantic-aware Alignment}
In the first stage, we utilize the WebVid-10M~\cite{bain2021frozen} dataset, which provides a large number of video-caption pairs, to train the periodicity transformer. A pre-trained CLIP text encoder~\cite{radford2021learning} embeds each video caption, producing the textual embedding $\mathcal{E}$. To bridge the multimodal gap, we align the periodic embedding $\mathcal{Z}$ with $\mathcal{E}$ by maximizing their mutual information, thereby achieving a text semantic-aware alignment.

Specifically, we define the similarity calculation as $g(\mathcal{Z}, \mathcal{E})$, where each of the $n$ query embeddings in $\mathcal{Z}$ is compared pairwise with $\mathcal{E}$, and the highest similarity is selected as the final result. Based on $g$, we compute the softmax-normalized video-to-text and text-to-video similarities for each periodic and textual embedding in a batch of size $K$ as follows:
\begin{gather}
s_k^\mathrm{v2t}(\mathcal{Z}) = \frac{\exp (g(\mathcal{Z},\mathcal{E}_k) / \tau)}{\sum_{k=1}^K \exp (g(\mathcal{Z},\mathcal{E}_k)/ \tau)},\\
s_k^\mathrm{t2v}(\mathcal{E}) = \frac{\exp (g(\mathcal{Z}_k, \mathcal{E})/ \tau)}{\sum_{k=1}^K \exp (g(\mathcal{Z}_k, \mathcal{E})/ \tau)},
\end{gather}
where $\tau$ is a temperature factor. Using the one-hot ground truths $y^\mathrm{v2t}(\mathcal{Z})$ and $y^\mathrm{t2v}(\mathcal{E})$, we define the training objective of video-text contrastive learning as follows:

\begin{equation}
\begin{aligned}
\mathcal{L}_\mathrm{vtc} = \frac{1}{K}\sum_{i=1}^K(\textbf{CE}(y^\mathrm{v2t}(\mathcal{Z}_i),s^\mathrm{v2t}(\mathcal{Z}_i))\\+\textbf{CE}(y^\mathrm{t2v}(\mathcal{E}_i),s^\mathrm{t2v}(\mathcal{E}_i))),
\end{aligned}
\end{equation}
where \textbf{CE} represents the Cross Entropy Loss.

\paragraph{Stage 2. Periodicity-aware Alignment}
Building on the semantic-aware alignment established in stage 1, the periodicity transformer gains an understanding of text. To enhance its understanding of periodicity, which is crucial for the RAC task, we replace the video captions with natural language description of periodicity, facilitating a smooth transition to periodicity-awareness.

We first encode this textual description, which encapsulates the semantic embeddings of periodicity, denoted as $\mathcal{E}_{per}$. Then, we employ periodicity-text contrastive learning to ensure alignment between periodic representations $\mathcal{Z}$ and periodicity semantics $\mathcal{E}_{per}$. In this setup, we use RAC datasets, and select video clips containing repetitive actions as positive samples and video clips without repetitive actions as negative samples. Our goal is to maximize the mutual information between positive samples and periodicity descriptions while minimizing it for negative samples. The periodicity-to-text similarity can be formulated as:
\begin{equation}
s^\mathrm{p2t}(\mathcal{Z}) = \sigma(g(\mathcal{Z},\mathcal{E}_{per})),
\end{equation}
where $\sigma$ represents the Sigmoid function. The ground truth $y^\mathrm{p2t}(\mathcal{Z})$ is 1 for positive samples and 0 for negative samples. Finally, the overall training objective of periodicity-text contrastive learning can be defined as:
\begin{equation}
\mathcal{L}_\mathrm{ptc} = \frac{1}{K}\sum_{i=1}^K\textbf{BCE}(y^\mathrm{p2t}(\mathcal{Z}_i),s^\mathrm{p2t}(\mathcal{Z}_i)),
\end{equation}
where \textbf{BCE} represents the Binary Cross Entropy Loss.

\paragraph{Stage 3. Instruction Tuning}
After the previous two stages, we conduct instruction tuning on the complete CountLLM to further align its responses with the given instructions. We apply LoRA~\cite{hu2021lora} to the pre-trained LLM and video encoder, leveraging a small set of learnable parameters for parameter-efficient adaptation to the RAC task. By combining the periodic tokens $\mathcal{P}_{\mathrm{token}}$ projected from $\Phi_P$ with the textual tokens $\mathcal{T}_{\mathrm{token}}$ from the text instruction, $\Phi_L$ predicts each output token $\{r_1, r_2, \dots, r_l\}$ and computes their softmax results for vocabulary classification $\{\mathcal{C}(r_1), \mathcal{C}(r_2), \dots, \mathcal{C}(r_l)\}$. Meanwhile, we tokenize the ground truth for each video clip into $\{y_1, y_2, \dots, y_l\}$ and define the training objective for instruction tuning as:
\begin{equation}
\mathcal{L}_\mathrm{llm}=\sum_{i} \textbf{CE}(\mathcal{C}(r_i), y_i).
\end{equation}

\section{Experiments}

\subsection{Experimental Setups}

\paragraph{Datasets}
We conduct experiments on three benchmarks: RepCount~\cite{hu2022transrac}, UCFRep~\cite{zhang2020context}, and Countix~\cite{dwibedi2020counting}. For RepCount and Countix, we tune the hyperparameters on validation set and report the results on test set, whereas for UCFRep, we directly report our results on validation set.

\paragraph{Evaluation Metrics}
We employ two widely used metrics in this task, which are \textbf{Off-By-One (OBO) count accuracy} and \textbf{Mean Absolute Error (MAE)}. OBO measures the accuracy rate of repetition count over the entire dataset, while MAE represents the normalized absolute error between the ground truth and the prediction. They can be defined as:
\begin{gather}
\mathrm{\bf{OBO}} = \frac{1}{N}\sum\limits_{i=1}^N[\vert \Tilde{c}_i-c_i\vert\leq 1]\\
\mathrm{\bf{MAE}} = \frac{1}{N}\sum\limits_{i=1}^N\frac{\vert \Tilde{c}_i-c_i\vert}{\Tilde{c_i}}
\end{gather}
where $\Tilde{c}_i$ denotes the ground truth, $c_i$ represents our prediction, and $N$ is the number of videos in the dataset.

\vspace{-0.2cm}

\paragraph{Baselines}
We compare with several baselines on RepCount and UCFRep, as listed in~\cref{tab:main_result1}. For Countix, we compare only with RepNet, Sight \& Sound, and ESCounts, as the other methods cannot effectively handle the lack of fine-grained annotations. For consistency, we report the performance of these baselines as stated in their respective papers. If results are unavailable, we re-implement them using the optimal hyperparameters from the original works.

\vspace{-0.2cm}

\paragraph{Implementation Details}
We implement two video encoders: Video Swin Transformer~\cite{liu2022video} and VideoMAE~\cite{wang2023videomae}, both pre-trained on Kinetics~\cite{carreira2019short}. Vicuna-7B~\cite{chiang2023vicuna} is used as the LLM. We set 64 learnable periodicity queries in the 12-layer periodicity transformer, and insert LoRA~\cite{hu2021lora} modules into the attention layers of both video encoder and LLM, with hidden dimension $r=16$. For training, we use WebVid-10M~\cite{bain2021frozen} in stage 1, and RAC datasets in stages 2 and 3. In stage 1, we train with 16-frame videos for 10 epochs, and for 50 epochs in stage 2; in stage 3, we use 32-frame videos for 50 epochs. For testing, video frames are sampled at intervals to ensure each clip contains 32 frames. Experiments are conducted on 8 NVIDIA A100 GPUs.

\begin{table}[t!]
    \centering
    \scalebox{0.8}{
    \begin{tabular}{l|c|cc|cc}
		\toprule
		\multirow{2}{*}{Method} &
        \multirow{2}{*}{Encoder} &
		\multicolumn{2}{c|}{RepCount} & 
        \multicolumn{2}{c}{UCFRep} \\
		& & MAE $\downarrow$ & OBO $\uparrow$ & MAE $\downarrow$ & OBO $\uparrow$ \\
		\midrule
        RepNet~\cite{dwibedi2020counting} & R2D50 & 0.995 & 0.013 & 0.998 & 0.009 \\
        Context~\cite{zhang2020context} & RX3D101 & 0.879 & 0.155 & 0.147 & 0.790 \\
        Si \& So$^{\text{\textdagger}}$~\cite{zhang2021repetitive} & R(2+1)D18 & 0.732 & 0.196 & \underline{0.143} & 0.800 \\
        TransRAC~\cite{hu2022transrac} & VSwinT & 0.443 & 0.291 & 0.441 & 0.430 \\
        Full~\cite{li2023full} & VSwinT & 0.410 & 0.327 & 0.461 & 0.333 \\
        MFL~\cite{li2024repetitive} & RX3D101 & 0.384 & 0.386 & 0.388 & 0.510 \\
        ESCounts~\cite{sinha2024every} & VideoMAE & 0.213 & \underline{0.563} & 0.216 & 0.704 \\
        \midrule
        \multirow{2}{*}{{\bf CountLLM}} & VSwinT & \underline{0.209} & 0.552 & 0.153 & \underline{0.802} \\
        & VideoMAE & \textbf{0.162} & \textbf{0.639} & \textbf{0.116} & \textbf{0.839} \\
		\bottomrule
    \end{tabular}
    }
    \vspace{-0.1cm}
    \caption{Comparisons on RepCount and UCFRep. The best results are highlighted in \textbf{bold}, and the second-best results are \underline{underlined}. $\text{\textdagger}$: Sight \& Sound. Our CountLLM surpasses previous approaches on two metrics, showcasing excellent counting accuracy.}
    \label{tab:main_result1}
    \vspace{-0.3cm}
\end{table}

\begin{table}[t!]
    \centering
    \scalebox{0.9}{
    \begin{tabular}{l|c|cc}
		\toprule
		\multirow{2}{*}{Method} &
        \multirow{2}{*}{Encoder} &
        \multicolumn{2}{c}{Countix} \\
		& & MAE $\downarrow$ & OBO $\uparrow$ \\
		\midrule
        RepNet~\cite{dwibedi2020counting} & R2D50 & 0.364 & 0.303 \\
        Sight \& Sound~\cite{zhang2021repetitive} & R(2+1)D18 & 0.307 & 0.511 \\
        ESCounts~\cite{sinha2024every} & VideoMAE & 0.276 & 0.673 \\
        \midrule
        \multirow{2}{*}{{\bf CountLLM}} & VSwinT & \underline{0.246} & \underline{0.706} \\
        & VideoMAE & \textbf{0.215} & \textbf{0.801} \\
		\bottomrule
    \end{tabular}
    }
    \vspace{-0.1cm}
    \caption{Comparisons on the Countix show that CountLLM demonstrates superior performance on two metrics.}
    \label{tab:main_result2}
\end{table}

\subsection{Main Results}
As shown in~\cref{tab:main_result1} and~\cref{tab:main_result2}, CountLLM outperforms previous methods across three benchmarks. On the RepCount dataset, with the same encoder, CountLLM surpasses TransRAC by $+0.261$ and Full by $+0.225$ on OBO metric. Using a more advanced encoder further boosts the OBO metric of CountLLM from $0.552$ to $0.639$. This advantage is also evident on the UCFRep and Countix datasets, where CountLLM achieves an OBO improvement of $+0.135$ over ESCounts on UCFRep and of $+0.128$ on Countix.

We analyze that previous approaches use handcrafted modules to regress periodic heatmaps, which limits their representational capacity. In contrast, CountLLM leverages pre-trained LLMs to incorporate powerful reasoning capabilities. Through explicit periodic text prompts and a carefully designed progressive training scheme, CountLLM enhances its periodicity awareness, enabling accurate capture of variable and non-homogeneous periodic patterns.

\begin{figure*}[t!]
  \includegraphics[width=\linewidth]{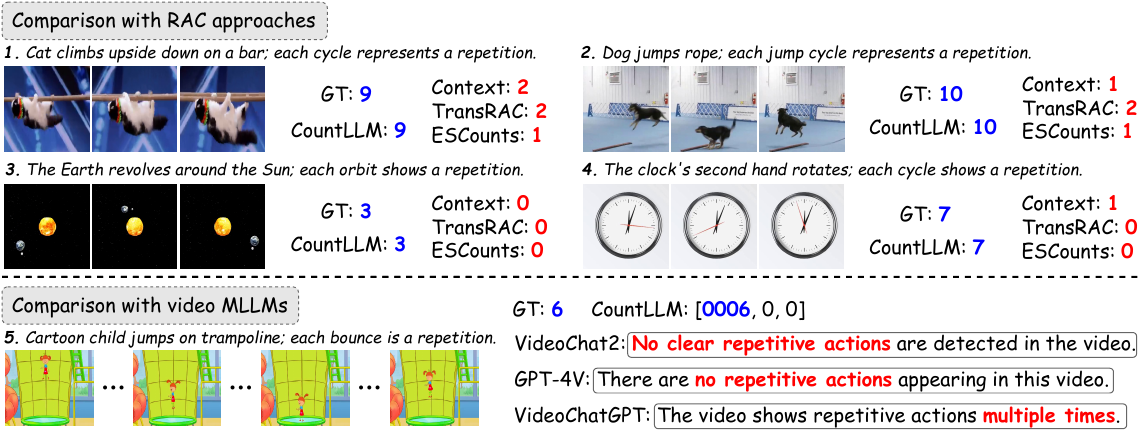}
  \caption{CountLLM demonstrates strong generalization when applied to \textbf{out-of-domain} repetitions, which differ significantly from the training data, rendering existing RAC approaches ineffective. Meanwhile, current video MLLMs lack periodicity awareness and perform poorly in counting them. For clearer understanding, the content of these videos is described in the figure (which is \textbf{not} input to CountLLM).}
  \label{fig:out-of-domain}
\end{figure*}

\begin{table}[t!]
    \centering
    \scalebox{0.9}{
    \begin{tabular}{l|cc|cc}
		\toprule
		\multirow{2}{*}{Method} &
		\multicolumn{2}{c|}{RepCount $\rightarrow$ UCFRep} &
        \multicolumn{2}{c}{RepCount $\rightarrow$ Countix} \\
		& MAE $\downarrow$ & OBO $\uparrow$ & MAE $\downarrow$ & OBO $\uparrow$ \\
		\midrule
        RepNet & 0.998 & 0.009 & 0.729 & 0.183 \\
        Context & 0.762 & 0.412 & 0.654 & 0.208\\
        TransRAC & 0.640 & 0.324 & 0.593 & 0.364 \\
        MFL$^{\text{\textdagger}}$ & 0.523 & 0.350 & --- & --- \\
        ESCounts & 0.317 & 0.571 & 0.374 & 0.521 \\
        \midrule
        \textbf{CountLLM} & \textbf{0.139} & \textbf{0.816} & \textbf{0.281} & \textbf{0.730} \\
		\bottomrule
    \end{tabular}
    }
    \caption{Comparison of cross-dataset generalization. $X \rightarrow Y$ denotes that the model is trained on $X$ and tested on $Y$. $\text{\textdagger}$: Results on Countix are unavailable as MFL is not open-source.}
    \label{tab:cross}
\end{table} 

\subsection{Cross-Dataset Generalization}
We compare the challenging cross-dataset generalization of different approaches, where each model is trained on RepCount and tested on UCFRep and Countix. As shown in~\cref{tab:cross}, CountLLM demonstrates excellent generalization across two datasets. \textbf{1) Horizontal comparison}: Compared with ESCounts, CountLLM achieves improvements of $+0.245$ on UCFRep and $+0.209$ on Countix, respectively. The performance gains over other approaches are even higher. \textbf{2) Vertical comparison}: The performance of all methods declines to some extent in cross-dataset setting (as compared with the regular setting in~\cref{tab:main_result1} and~\cref{tab:main_result2}). However, CountLLM shows only a slight decline, demonstrating stronger robustness than other methods.

Overall, both horizontal and vertical comparisons indicate that our CountLLM effectively leverages general periodic semantics in text prompts and the reasoning capabilities of LLM, thereby reducing overfitting to the training set and enhancing generalization in the RAC task.

\begin{table}[t!]
    \centering
    \scalebox{1.0}{
    \begin{tabular}{l|c|c|c}
    \toprule
    Methods & Encoder & MAE $\downarrow$ & OBO $\uparrow$ \\
    \midrule
    RepNet~\cite{dwibedi2020counting} & R2D50 & 0.86 & 0.35 \\
    OVRCounter~\cite{dwibedi2024ovr} & ViViT~\cite{arnab2021vivit} & 0.39 & 0.59 \\
    \textbf{CountLLM} & VSwinT & \textbf{0.26} & \textbf{0.65} \\
    \bottomrule
    \end{tabular}
    }
    \caption{Comparison of generalization on OVR-Kinetics dataset.}
    \label{tab:ovr}
\end{table} 

\subsection{Out-of-Domain Generalization}
We compare the generalization of CountLLM with several RAC approaches and video MLLMs by selecting various out-of-domain repetitions, including repetitive actions of animals and cartoon characters, as well as periodic motions of non-living entities, such as clock movements and planetary motions. The results are illustrated in~\cref{fig:out-of-domain}.
\textbf{1) RAC approaches.} Existing RAC methods fail to accurately count these out-of-domain actions. We analyze that they primarily rely on supervised learning from specific training sets containing only common human fitness activities, which makes them prone to overfitting and limits their ability to understand general periodic patterns.
\textbf{2) Video MLLMs.} Current video MLLMs, such as VideoChat2~\cite{li2024mvbench}, GPT-4V~\cite{gpt4v}, and VideoChatGPT~\cite{maaz2023video}, cannot reliably count these various repetitions. We attribute this limitation to their lack of periodicity awareness. These MLLMs focus on general video understanding during training, thus lacking specific periodic guidance and struggling with counting these challenging periodic motions.

In contrast, our CountLLM accurately counts repetitions that differ entirely from those in the training set. We analyze that the explicit periodic text prompt encapsulates the key properties of repetitive actions, which are independent of specific action categories, thus preventing overfitting to the training set. Meanwhile, through instruction fine-tuning tailored for RAC task, the LLM develops periodicity awareness, enhancing its ability to understand and extract periodic patterns. These design choices enable CountLLM to effectively handle unseen and out-of-domain actions.

Additionally, we test the generalization on the recently proposed OVR dataset~\cite{dwibedi2024ovr}, which contains a large number of videos with a huge variety of open vocabulary actions. As shown in~\cref{tab:ovr}, our CountLLM surpasses the open vocabulary counting baselines, which further demonstrates the strong generalizability of CountLLM.

\subsection{Ablation Studies}

\paragraph{Instruction Conversation}
We analyze the components in instruction-driven conversation: \textbf{1) W/o periodicity description.} In this case, we remove the periodicity description and directly question the LLM. As shown in~\cref{tab:instruction}, the lack of textual description leads to a decline in both metrics. This occurs because the LLM cannot fully grasp the definition of repetitive actions before performing the RAC task. \textbf{2) W/o decimal string answers.} Here, we remove the fixed-length decimal string answers and instead use additional learnable tokens to represent counting values. Specifically, we add a set of learnable tokens $\langle 0000 \rangle$, $\langle 0001 \rangle$, ..., $\langle 9999 \rangle$ into the tokenizer to represent each value and tune them during training. As illustrated in~\cref{tab:instruction}, these learnable tokens result in a notable decline in performance. We attribute this to the fact that, compared with decimal strings, learnable tokens lack flexibility and require retraining of the embedding layer, with their effectiveness constrained by the limited scale of current RAC datasets.

\begin{table}[t!]
    \centering
    \scalebox{1.0}{
    \begin{tabular}{l|cc}
		\toprule
		\multirow{2}{*}{Settings} &
		\multicolumn{2}{c}{RepCount Val Set}\\
		& MAE $\downarrow$ & OBO $\uparrow$ \\
		\midrule
        w/o periodicity description & 0.305 & 0.476 \\
        w/o decimal string answers & 0.392 & 0.387 \\
        \midrule
        All (our CountLLM) & \textbf{0.166} & \textbf{0.631} \\
		\bottomrule
    \end{tabular}
    }
    \caption{Ablation studies on instruction conversation. \textbf{W/o periodicity description}: We directly question the LLM without providing a description. \textbf{W/o decimal string answers}: We adjust the answer format by introducing additional learnable tokens.}
    \label{tab:instruction}
\end{table}

\begin{table}[t!]
    \centering
    \scalebox{1.0}{
    \begin{tabular}{cc|cc}
		\toprule
		\multicolumn{2}{c|}{Progressive Training Stages} &
		\multicolumn{2}{c}{RepCount Val Set}\\
		S-A Alignment & P-A Alignment & MAE $\downarrow$ & OBO $\uparrow$ \\
		\midrule
        & & 0.402 & 0.368 \\
        \checkmark & & 0.363 & 0.417 \\
        & \checkmark & 0.236 & 0.551 \\
        \checkmark & \checkmark & \textbf{0.166} & \textbf{0.631} \\
		\bottomrule
    \end{tabular}
    }
    \caption{Ablation studies on the semantic-aware (\textbf{S-A}) alignment and periodicity-aware (\textbf{P-A}) alignment in progressive training.}
    \label{tab:training}
\end{table}

\paragraph{Progressive Training}
We propose semantic-aware and periodicity-aware alignment to optimize the periodicity transformer before instruction tuning, with each stage analyzed in~\cref{tab:training}. \textbf{1)} If both semantic-aware and periodicity-aware alignment are removed, the metrics decline significantly, with the OBO metric decreasing by $0.263$. \textbf{2)} When we reintroduce the semantic-aware alignment, the performance improves slightly, which we attribute to the partial bridging of the gap between video and text modalities. \textbf{3)} When we apply only the periodicity-aware alignment, the performance shows a substantial improvement over the baseline, with the OBO metric increasing from $0.368$ to $0.551$. We conclude that textual periodicity description encapsulates essential periodic semantics that are valuable in the RAC task. By aligning with periodicity, robust periodic representations are captured, providing the LLM with useful context. \textbf{4)} Finally, when we progressively apply both semantic-aware and periodicity-aware alignment, the MAE and OBO metrics reach their highest values.

\begin{table}[t!]
    \centering
    \scalebox{1.0}{
    \begin{tabular}{cc|cc}
		\toprule
		\multicolumn{2}{c|}{LoRA Adaptation} &
		\multicolumn{2}{c}{RepCount Val Set}\\
		$\Phi_V(\cdot)$ & $\Phi_L(\cdot)$ & MAE $\downarrow$ & OBO $\uparrow$ \\
		\midrule
        \multicolumn{4}{l}{\emph{w/o Periodicity Transformer}} \\
        \midrule
        \checkmark & \checkmark & 0.498 & 0.260 \\
        \midrule
        \multicolumn{4}{l}{\emph{w Periodicity Transformer}} \\
        \midrule
        & & 0.409 & 0.357 \\
        \checkmark & & 0.239 & 0.546 \\
        & \checkmark & 0.317 & 0.498 \\
        \checkmark & \checkmark & \textbf{0.166} & \textbf{0.631} \\
		\bottomrule
    \end{tabular}
    }
    \caption{Ablation studies on the model designs. \textbf{W/o Periodicity Transformer}: Video features are projected using only a linear layer, with both S-A and P-A alignments removed. \textbf{W Periodicity Transformer}: Both S-A and P-A alignments are included.}
    \label{tab:model_design}
\end{table}

\paragraph{Model Designs}
We investigate several model designs in CountLLM: \textbf{1) Periodicity transformer.} To establish a baseline, we remove the periodicity transformer and instead use a linear projector to directly convert video features into LLM tokens. As shown in~\cref{tab:model_design}, this design results in a substantial performance decline, underscoring the importance of the periodicity transformer in compressing redundant video features into essential periodic representations. \textbf{2) LoRA adaptation.} We apply parameter-efficient LoRA during instruction tuning to improve the adaptation of the pre-trained $\Phi_{V}$ and $\Phi_{L}$. As illustrated in~\cref{tab:model_design}, freezing both $\Phi_{V}$ and $\Phi_{L}$ limits performance, as they may not fully adapt to the RAC datasets. In contrast, inserting learnable LoRA layers into either $\Phi_{V}$ or $\Phi_{L}$ markedly enhances both metrics. Finally, fine-tuning both $\Phi_{V}$ and $\Phi_{L}$ allows CountLLM to achieve optimal performance.
\section{Conclusion}
CountLLM advances repetitive action counting by leveraging the powerful reasoning capabilities of pre-trained LLMs through explicit periodic text prompts. By incorporating a structured template for periodicity-based instruction and a progressive training scheme, CountLLM achieves superior performance across multiple benchmarks and exhibits robust generalization to out-of-domain action scenarios. These results underscore the potential of large language models in opening promising avenues for broader applications in video analysis and beyond.

While CountLLM achieves impressive accuracy in repetitive action counting, it is partially constrained by the computational demands of large language models, which require substantial GPU resources. This limitation may be alleviated by using lightweight LLMs. Furthermore, although our structured, periodicity-focused templates enhance the model's interpretive capabilities, accuracy could potentially be improved with even more precise periodicity descriptions. Future work may explore optimized in-context learning strategies, potentially incorporating targeted examples to strengthen periodicity recognition.

\section*{Acknowledgement}
This work was supported in part by the Pioneer Centre for AI, DNRF grant number P1.

{
    \small
    \bibliographystyle{ieeenat_fullname}
    \bibliography{main}
}


\end{document}